\documentclass[letterpaper, 10 pt, conference]{ieeeconf}  %

\usepackage{amsmath}
\usepackage{amssymb}
\usepackage{bm}
\usepackage{booktabs}
\usepackage{caption}
\usepackage{cite}
\usepackage{colortbl}
\usepackage{hyperref}
\usepackage{mathptmx} %
\usepackage{makecell}
\usepackage{setspace} %
\usepackage{subcaption}
\usepackage{xcolor}
\usepackage{tikz}

\usepackage[nameinlink]{cleveref} %

\definecolor{uoftblue}{RGB}{30, 55, 101} %

\definecolor{uoftsecondaryblue}{RGB}{0,127,163} %

\definecolor{uoftpurple}{RGB}{109,36,122} %
\definecolor{uoftwarmred}{RGB}{220,70,51} %
\definecolor{uoftcoolblue}{RGB}{111,199,234} %
\definecolor{uoftteal}{RGB}{0,161,137} %
\definecolor{uoftfuchsia}{RGB}{171,19,104} %
\definecolor{uoftdarkgreen}{RGB}{13,83,77} %
\definecolor{uoftyellow}{RGB}{241,197,0} %
\definecolor{uoftlightgreen}{RGB}{141,191,46} %

\definecolor{uoftcoolgray}{RGB}{208,209,201} %

\hypersetup{
    colorlinks=true,
    allcolors=uoftsecondaryblue,
    pagebackref
}

\Crefname{figure}{Fig.}{Figs.}
\crefname{figure}{Fig.}{Figs.}
\Crefname{appendix}{Appendix}{Appendices}
\crefname{appendix}{Appendix}{Appendices}
\Crefname{table}{Table}{Tables}
\crefname{table}{Table}{Tables}
\Crefname{section}{Section}{Section}
\crefname{section}{Section}{Section}
\Crefname{equation}{Eq.}{Eqs.}
\crefname{equation}{Eq.}{Eqs.}

\IEEEoverridecommandlockouts                              %

\title{\LARGE \bf
        Dynamics Distillation for Efficient and Transferable Control Learning
}

\author{Xunjiang Gu$^{1,4}$, Kashyap Chitta$^{2}$, Mahsa Golchoubian$^{1,4}$, Vladimir Suplin$^{3}$, Igor Gilitschenski$^{1,4}$%
\thanks{$^{1}$University of Toronto. \texttt{\small alfred.gu@mail.utoronto.ca, mahsa.golchoubian@utoronto.ca, gilitschenski@cs.toronto.edu}}%
\thanks{$^{2}$NVIDIA. \texttt{\small kchitta@nvidia.com}}%
\thanks{$^{3}$General Motors. \texttt{\small vladimir.suplin@gm.com}}%
\thanks{$^{4}$Vector Institute.}%
}

\begin{document}

\maketitle
\thispagestyle{empty}
\pagestyle{empty}

\begin{abstract}
Robust control policy learning for autonomous driving requires training environments to be both physically realistic and computationally scalable, properties that existing simulators provide only in isolation. 
We introduce Sim2Sim2Sim{%
\renewcommand{\thefootnote}{$*$}%
\footnote{Code: \href{https://github.com/alfredgu001324/Sim2Sim2Sim}{https://github.com/alfredgu001324/Sim2Sim2Sim}}%
}, a framework that bridges high-fidelity vehicle simulation and scalable reinforcement learning (RL) by distilling simulator dynamics into a highly parallelizable learned dynamics model. 
By training control policies purely within this distilled environment and deploying them back into the high-fidelity source simulator, we demonstrate more efficient policy optimization and reliable transfer under challenging dynamics. 
We further show that predictive accuracy alone does not fully characterize a learned dynamics model's suitability as an RL training environment, which should also be assessed by the quality of the policies it enables.
\end{abstract}

\section{INTRODUCTION}
\label{sec:intro}

Handling abrupt changes in vehicle dynamics remains challenging in autonomous driving, such as during transitions between dry asphalt and ice.
Whether the upstream planner is rule-based~\cite{fan2018baidu}, end-to-end~\cite{Weng2024PARADrive}, or a vision-language-action model~\cite{wang2025alpamayo}, every autonomous driving stack ultimately relies on a low-level controller to translate planned intent into physical motion. 
Even a perfectly planned trajectory does not guarantee system safety if the controller cannot reliably execute it under sudden dynamics shifts.
In such regimes, small perturbations can quickly amplify into large deviations or even complete control failure, making it essential to train control policies that are robust to challenging conditions.

Reinforcement learning provides a flexible framework for handling such nonlinear and time-varying behavior, but its effectiveness is fundamentally constrained by the environments in which policies are trained. 
Physics-accurate simulators capture rich vehicle dynamics and realistic failure modes but are prohibitively expensive for large-scale RL training, whereas fast, highly parallelized simulators enable efficient policy optimization but rely on simplified dynamics that fail to reflect real vehicle behavior under challenging conditions. 
Neither alone is sufficient for learning scalable and reliable policies under realistic dynamics, motivating a paradigm that reconciles these complementary properties.

\begin{figure}[t]
\centering
\includegraphics[width=\linewidth]{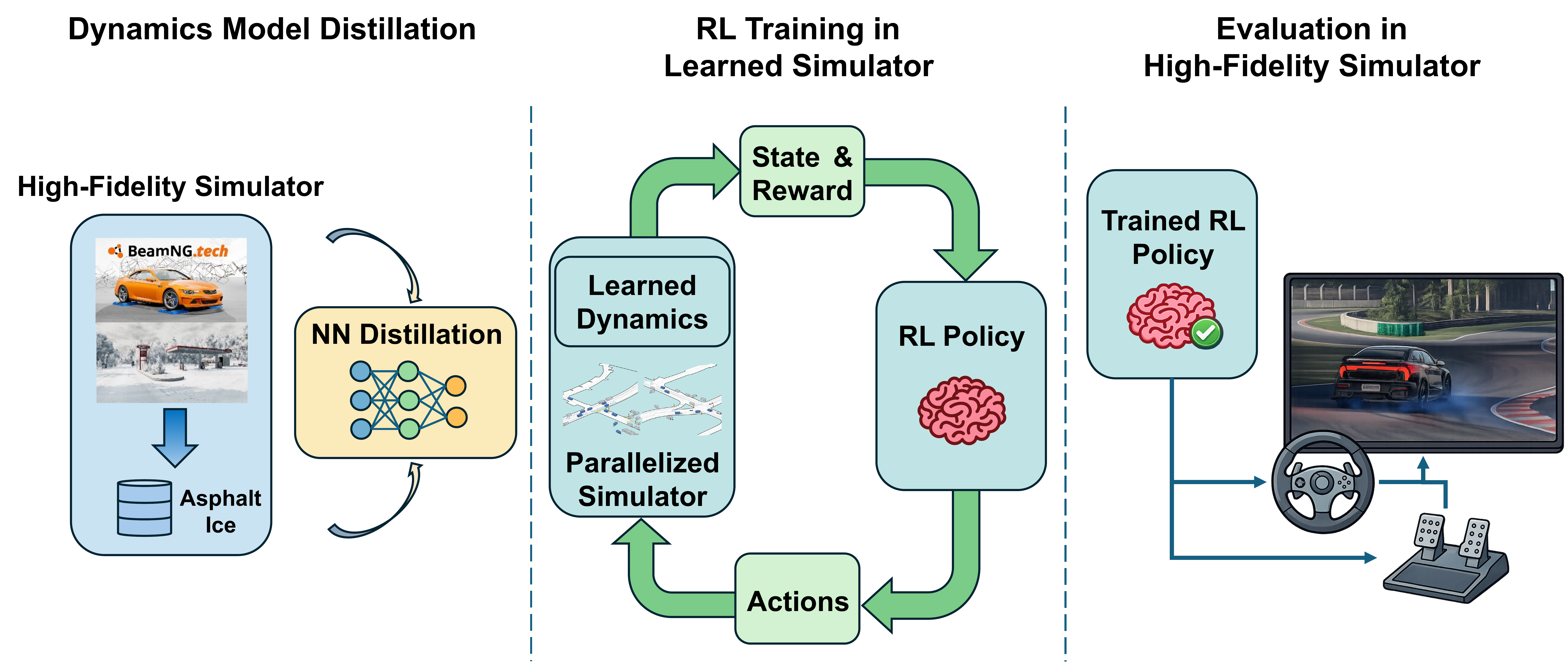}
\caption[Sim2Sim2Sim training pipeline]{The Sim2Sim2Sim framework operates in three stages: (1) dynamics distillation from a physics-accurate simulator into learned transition models, (2) massively parallel policy optimization using the distilled dynamics as a fast simulation backend, and (3) zero-shot deployment back to the high-fidelity simulator for closed-loop evaluation under challenging conditions including abrupt surface transitions.}
\label{fig:hero}
\vspace{-0.3cm}
\end{figure}

In this work, we introduce Sim2Sim2Sim (\cref{fig:hero}), a training framework that bridges high-fidelity vehicle simulation and scalable reinforcement learning through dynamics distillation. 
Instead of training control policies directly in expensive physics-accurate simulators, we first distill their vehicle dynamics into a learned transition model that can be embedded into a separate highly parallelized simulation backend. 
Control policies are then trained entirely within this learned environment and subsequently deployed back into the original high-fidelity environment for closed-loop evaluation. 
By separating where dynamics are learned, where policies are optimized, and where policies are evaluated, Sim2Sim2Sim demonstrates efficient large-scale training while preserving the physical characteristics required for reliable transfer.

\textbf{Contributions.} Our core contributions are threefold: First, we propose Sim2Sim2Sim, a dynamics distillation framework that enables efficient large-scale RL policy training by embedding high-fidelity simulator dynamics into a fast, parallelized simulation backend. Second, we demonstrate reliable zero-shot transfer of policies trained on distilled dynamics to high-fidelity simulation under abrupt friction transitions, enabled by grounding policy learning in real-world driving trajectories with controllable surface variation. Finally, we introduce a closed-loop evaluation protocol for learned dynamics models that reveals failure modes and robustness characteristics invisible to standard open-loop prediction metrics.

\section{Related Work}
\label{sec:literature_review}

\textbf{Dynamics Model Learning.} Vehicle dynamics modeling spans physics-based, data-driven, and hybrid formulations. 
Classical models range from point-mass to single-track and multi-body representations, trading computational efficiency for physical fidelity~\cite{Althoff2017CommonRoad,Kapania2016,Timings2013}, though their accuracy degrades under changing conditions such as tire wear or surface variation~\cite{Vicente2021}. 
Parameter learning and system identification methods address this by estimating physics model coefficients from data, using physics-informed neural networks and Gaussian processes to identify drivetrain and tire parameters~\cite{Kabzan2019,Xu2022,Chrosniak2024DDM}, but remain constrained by the expressiveness of the underlying physics structure.
Purely data-driven models bypass these limitations by learning state transitions directly from observations and control inputs~\cite{Spielberg2019,Hermansdorfer2020,Punjani2015}, capturing complex nonlinear dynamics at the cost of generalization and interpretability. 
Hybrid residual approaches combine physics-based models with learned corrections~\cite{Ning2023Scalable, Baier2021HybridPA}, retaining the physical structure while improving accuracy, though performance is limited by the base model's fidelity. 
Our work leverages these paradigms but shifts focus from single-step prediction accuracy to how different dynamics models support scalable and transferable RL training.

\textbf{Simulation in Autonomous Driving.}
Simulators for autonomous driving can be broadly categorized into perception-focused, parallelism-focused, and physics-focused systems. 
Perception-focused simulators provide realistic sensor observations but incur high computational cost and limited scalability~\cite{Dosovitskiy2017CORL, Cao2025CORL}; our work focuses on control training environments and relates primarily to the latter two categories.
Parallelism-focused simulators operate at the trajectory level using simplified vehicle models, enabling large-scale reinforcement learning~\cite{cusumano2025gigaflow, Gulino2023Waymax,Li2022Metadrive}. 
One example is GPUDrive~\cite{Kazemkhani2024GPUDrive}, built on top of the Madrona engine~\cite{shacklett23madrona}, enables GPU-native simulation across hundreds of parallel environments initialized from the Waymo Open Motion Dataset~\cite{waymo_open_motion_dataset}, consistent with findings that data volume and interaction diversity are critical for learning better policies~\cite{hwang2024emma, naumann2025data}.
However, their reliance on kinematic dynamics limits suitability for realistic control validation.
Physics-focused simulators—such as BeamNG.tech~\cite{beamng_tech}, Assetto Corsa Gym~\cite{Remonda2024Assetto}, and Gran Turismo~\cite{Wurman2022Outracing}—provide high-fidelity vehicle dynamics including nonlinear tire–road interaction and load transfer, but training RL policies directly within them remains expensive and difficult to scale~\cite{Jaeger2025CoRL, Grooten2025SPARC}. 
Our work bridges this gap by distilling high-fidelity dynamics into a learned transition model for scalable policy training, then deploying back to the original environment for closed-loop evaluation.
Our approach is similar in spirit to robotic world models (RWM)~\cite{Li2025RWM}, but differs in two key aspects: we retain an explicit GPUDrive\cite{Kazemkhani2024GPUDrive} front-end for collision and off-road checking, and ground policy learning in large-scale real-world driving trajectories to ensure compatibility with real-world motion distributions.

\textbf{Low-Level Control for Autonomous Systems.}
Autonomous driving typically follows a hierarchical framework where a planner generates a reference trajectory tracked by a low-level controller. 
Prior work improves tracking performance by incorporating dynamics constraints, friction limits, or uncertainty into Model Predictive Control (MPC) formulations~\cite{Mihalkov2024, Hu2023, Gao2025}, but these approaches rely on precise system identification and repeated online optimization, making them computationally expensive and difficult to scale across varying dynamics.
Reinforcement learning has emerged as a complementary alternative for low-level autonomous driving control, offering much faster inference through a single forward pass and the ability to handle complex, nonlinear behaviors~\cite{Wurman2022Outracing, cai2020high, chen2024deep}, but it lacks explicit mechanisms for enforcing physical constraints at deployment. 
Contextual and adaptive RL methods address this by conditioning on environment parameters or latent context, with related approaches explored in robotics for quadrotor control~\cite{Eschmann2025RAPTOR} and legged locomotion~\cite{Liu2025LocoFormer}. 
In autonomous racing, SPARC~\cite{Grooten2025SPARC} trains adaptive controllers directly in Gran Turismo to generalize across unseen vehicles, but relies on policy optimization within a high-fidelity simulator, incurring long training times, and is limited to single-agent racing with track-specific observations.

\section{Dynamics Model Distillation}
\label{sec:dynamics}

In our Sim2Sim2Sim framework, BeamNG.tech~\cite{beamng_tech} serves as the high-fidelity source simulator from which vehicle dynamics are distilled into a learned backend, enabling large-scale RL-based control training within GPUDrive~\cite{Kazemkhani2024GPUDrive}. 
We instantiate several dynamics model families spanning different levels of physical priors and representational capacities to study how dynamics fidelity, structural bias, and surface awareness influence downstream policy learning.

\subsection{Problem Formulation}
At timestep $t$, the vehicle state is $s_t = [x_t, y_t, \phi_t, v_{x,t}, v_{y,t}, \omega_t]$ where $(x,y)$ are position, $\phi$ is heading, $v_x,v_y$ are longitudinal and lateral velocities, and $\omega$ is yaw rate. The control input is $a_t = [\tau_t, \delta_t]$, where $\tau_t$ and $\delta_t$ are BeamNG's normalized throttle/brake and steering commands. 
We represent motion using incremental body-frame deltas $\Delta s^{(b)}_t$, where the superscript $(b)$ denotes the vehicle body frame, enabling the dynamics model to capture incremental motion independent of absolute position and heading. 
Surface labels $\sigma_t \in \{\mathrm{asphalt}, \mathrm{ice}\}$ are optionally included.

Each model receives a short history window $h_t = \big\{ \Delta s^{(b)}_{t-H+1:t},\; a_{t-H+1:t} \big\}$ and follows the single-step formulation:
\begin{equation}
    \hat{\Delta s}^{(b)}_{t+1} = f_\theta(h_t, a_t \, ; \, \sigma_t),
    \label{eq:dynamics}
\end{equation}
where $\sigma_t$ is provided only for models that explicitly condition on surface labels. 
All models are trained with an MSE loss and rolled out autoregressively during RL training, making closed-loop stability an important property.

\subsection{Model Families}

We evaluate four dynamics model families, adapting their core design principles to fit within a unified input--output interface while retaining each approach's defining structural assumptions. 
All models take normalized BeamNG control commands as input for direct deployment compatibility.

\subsubsection{Kinematic Model with Learned Actuation}
A kinematic single-track model combined with two MLPs that map normalized throttle and steering commands to longitudinal acceleration and steering rate. 
The model enforces a no-slip constraint, making it lightweight but limited under aggressive maneuvers or low-friction surfaces, serving as a structured low-fidelity baseline.

\subsubsection{Physics-Constrained Deep Dynamics Model (DDM)}
The DDM~\cite{Chrosniak2024DDM} is based on a dynamic single-track formulation with drivetrain forces and Pacejka tire models. 
A recurrent encoder estimates uncertain physical parameters bounded through a Physics Guard mechanism to ensure physically plausible values, introducing strong physical structure while retaining flexibility across varying surface conditions.

\subsubsection{Transformer Dynamics Model}
A fully data-driven sequence model~\cite{Xiao2025Anycar} that predicts state deltas by attending to recent state-action history. 
A surface-conditioned variant augments the state embedding with learned surface representations, capturing temporal dependencies without imposing explicit physical constraints.

\subsubsection{Residual Learning for Correction}
A hybrid model~\cite{Miao2025DYTR} that augments a physics-based base model with a Transformer-based residual network conditioned on state--action history and optional surface labels, combining structural priors with the flexibility of data-driven correction. 
This residual formulation is model-agnostic; in our framework, we instantiate it on top of the DDM as the base model since it provides the strongest physical inductive bias among the evaluated model families. 

\section{RL Policy Learning}
\label{sec:policy_learning}

Given a learned dynamics model $f_\theta$, we train a trajectory-tracking control policy by integrating the model as the transition operator inside a fast vectorized simulation backend. 
At each timestep, the policy outputs normalized control commands passed to the dynamics model, which predicts the next body-frame state delta used to update the global state. 
Although we use GPUDrive in our experiments, this approach is compatible with any accelerated simulation framework that supports custom transition functions.

\subsection{Simulation Backend Integration}

The learned dynamics model predicts a body-frame delta as defined in \cref{eq:dynamics}, where $h_t$ is the recent state--action history and $\sigma_t$ is the optional surface label. 
To expose the policy to friction variations, $\sigma_t$ can be switched according to a stochastic schedule during training, and can also be included in the policy observation to construct oracle policies. 
At evaluation time in GPUDrive, $\sigma_t$ can be set to any user-defined surface sequence for controlled evaluation.

\subsection{MDP Formulation}

We formulate the control problem as a finite-horizon MDP $\mathcal{M} = (\mathcal{S}, \mathcal{A}, f_\theta, \mathcal{R}, \rho_0, \gamma, T)$, where $\mathcal{S}$ and $\mathcal{A}$ are the state and action spaces, $f_\theta$ is the learned transition operator, $\mathcal{R}$ is the reward function, $\rho_0$ is the initial state distribution, $\gamma$ is the discount factor, and $T$ is the episode horizon. We identify the state with the policy observation $s_t := o_t$. 
The observation $o_t$ consists of an ego-state vector containing normalized body-frame velocities, yaw rate, and vehicle dimensions, together with a sequence of $K=10$ planned trajectory deltas
\begin{equation}
    \Delta w_{t,i} = (\Delta x_i, \Delta y_i, \Delta \phi_i, \Delta v_i), 
    \quad i = 1,\dots,K,
\end{equation}
transformed into the ego frame following common conventions in trajectory prediction~\cite{gu2024producing}. The continuous action space is $a_t = [\tau_t, \delta_t] \in [-1,1]^2$.

\subsection{Policy Architecture}

The control policy $\pi_\phi(a_t \mid o_t)$ is implemented as a feedforward actor--critic network. 
Ego-state features are encoded using an MLP, while planned trajectory waypoints are projected into a latent space, augmented with learned positional encodings, and processed by a Transformer encoder. 
The resulting trajectory embedding is mean-pooled and concatenated with the ego-state embedding to form a shared backbone representation. 
The actor head outputs the mean and state-dependent log standard deviation of a diagonal Gaussian distribution squashed via $\tanh$, and the critic head predicts a scalar value estimate.

\subsection{Reward}

The reward at timestep $t$ follows a multiplicative progress-penalty formulation~\cite{Jaeger2025CoRL}, defined as
\begin{equation}
    r_t = \Delta P_t \cdot \prod_i \exp(-\beta_i e_{t,i})
          \;-\; \mathbf{1}[\text{termination}],
\end{equation}
where $\Delta P_t$ denotes incremental progress along the reference trajectory and $\{e_{t,i}\}$ are soft error terms including cross-track error, heading deviation, speed deviation, a jerk penalty, and an action rate penalty, each weighted by $\beta_i$. 
Tracking errors are computed by comparing $s_{t+1}$ to the first waypoint of the planned trajectory provided at timestep $t$. 
Episodes terminate immediately upon collision or off-road events, incurring a fixed negative reward.

\section{Experiments}
\label{sec:expt}

\subsection{Simulation Environments}
\label{sec:env_data}
We evaluate our Sim2Sim2Sim framework using two complementary simulation environments: BeamNG.tech~\cite{beamng_tech}, which serves as a high-fidelity source simulator and final validation platform, and GPUDrive~\cite{Kazemkhani2024GPUDrive}, which provides a fast, scalable backend for reinforcement learning.

\textbf{BeamNG.tech} is a high-fidelity driving simulator built on a soft-body physics engine~\cite{beamng_tech}, producing complex traction behavior under varying surface conditions that makes it suitable both for collecting realistic vehicle dynamics data and for final closed-loop validations.

\textbf{GPUDrive} is a GPU-native, high-throughput multi-agent driving simulator designed for large-scale reinforcement learning~\cite{Kazemkhani2024GPUDrive}. 
In our framework, learned dynamics models replace the native kinematic update, enabling scalable policy learning while preserving dynamics characteristics distilled from BeamNG.tech, with all dynamics updates executed in a single batched GPU operation.

\subsection{Datasets and Policy Training Scenarios}
\label{sec:datasets}
\textbf{Dynamics Dataset.}
We collected 8 hours of driving data in BeamNG.tech across two surface conditions: 4 hours on asphalt and 4 hours on mixed surfaces that include friction transitions between asphalt and ice. 
A human driver performs diverse maneuvers including straight driving, turning, U-turns, highway driving, and abrupt braking. 
Data is recorded at 200~Hz and subsampled to 10~Hz to match the policy training frequency.
After preprocessing with a 12-step (1.2~s) history horizon, the dataset contains 293{,}148 samples, split 80\%/20\% for training and validation.

\textbf{Policy Training Datasets.}
Policies are trained on scenarios from the Waymo Open Motion Dataset (WOMD)~\cite{waymo_open_motion_dataset}, which contains over 100{,}000 real-world driving scenes with multi-agent logged trajectories and detailed road geometry, providing a rich distribution of reference motions for control learning. 
The planned trajectory, which forms part of the policy observation during training, is taken directly from these logs. 
For evaluation, we additionally incorporate the Waymo End-to-End (E2E) dataset~\cite{xu2025wod}, yielding 84{,}757 scenarios that include challenging behaviors such as construction zones, pedestrian and cyclist interactions, lane changes, and cut-ins. 
All scenarios in GPUDrive operate at 10~Hz.

\subsection{Evaluation Protocol}
\label{sec:eval_protocol}
We evaluate our framework at three levels: (i) open-loop evaluation of learned dynamics models, (ii) closed-loop policy performance inside GPUDrive, and (iii) Sim2Sim2Sim transfer evaluation in BeamNG. 
This separation allows us to disentangle dynamics prediction accuracy, policy learning behavior, and real-world transfer robustness.

\subsubsection{Dynamics Model Evaluation}
For open-loop evaluation, dynamics models are trained and evaluated on the asphalt subset following standard practice, using both single-step and short-horizon autoregressive rollout to characterize predictive behavior under nominal conditions. 
Surface-conditional models are additionally trained on the full dataset including mixed surfaces, but this variant is used solely as the dynamics backend for friction-aware policy training and is not evaluated in open-loop.

\subsubsection{Policy Evaluation in GPUDrive}
Closed-loop policy evaluation in GPUDrive is performed on two sets of real-world trajectories: the WOMD Mini validation split (150 scenes) and a subset of the Waymo E2E validation data from turning clusters (200 scenes). 
Observations are constructed using a lookahead-based reference trajectory extracted from logged data at each timestep.
To assess robustness to surface variability, we train policies under different surface exposure regimes and evaluate them under both matched and mismatched surface conditions.

\subsubsection{Sim2Sim Transfer Evaluation in BeamNG}
We evaluate zero-shot policy transfer on the Putnam Park track~\cite{Kulkarni2023} (total length 2.765~km) under two settings, without any fine-tuning.
The first assesses tracking accuracy under nominal asphalt condition. 
The second introduces localized ice patches along the track, testing whether policies exposed to ice during training in distilled environment can transfer robustly to high-fidelity dynamics to handle friction transitions. 

\subsection{Metrics}
\label{sec:metrics}

\subsubsection{Dynamics Model Evaluation}
We report Single-Step Displacement Error (SSDE), Average Displacement Error over 10 steps (ADE@10), and Final Displacement Error after 10 steps (FDE@10), all measured in meters, to capture both single-step accuracy and short-horizon autoregressive rollout behavior.

\subsubsection{Policy Evaluation in GPUDrive}
We report both tracking-level metrics and scenario-level metrics. 
Tracking performance is measured at two levels. 
At the trajectory level, we report cross-track error (CTE, meters), Average Displacement Error (ADE, meters), and Final Displacement Error (FDE, meters). 
At the control level, we report position tracking error (PTE, meters), heading error (HE, radians), and speed tracking error (STE, m/s), each computed with one-step delay by comparing $s_{t+1}$ to the first waypoint of the planned trajectory provided at timestep $t$. 
At the scenario level, we report success rate, collision rate, and off-road rate, each expressed as a percentage of agents averaged across all scenarios.

\subsubsection{Sim2Sim2Sim Transfer Evaluation in BeamNG}
We report CTE, PTE, HE, and STE computed analogously to the GPUDrive evaluation. We additionally report distance traveled before loss of control (as a percentage of total track length), policy inference latency (ms), and steering rate (Steer D1). Steer D1 is defined as the mean absolute difference between consecutive normalized steering commands $\delta_t \in [-1, 1]$, measuring control smoothness.

\subsection{Policies to Compare}
\label{sec:baselines}
We evaluate control policies trained using different dynamics model families as the simulation backend within our Sim2Sim2Sim framework, described in \cref{sec:dynamics}. Unless otherwise stated, policies use full body-frame velocity observations $(v_x, v_y, \omega)$ by default, with speed-only variants included as ablations.
For surface-conditional models, we denote the conditional Transformer as Trans C, and the DDM augmented with residual learning as DDM + Re-C, with the suffix -C indicating surface conditioning.
Policies targeting nominal asphalt conditions use a dynamics model trained exclusively on asphalt data, while policies targeting abrupt friction transitions use a dynamics model trained on the full dataset with $\sigma_t$ switched via a stochastic schedule during training.
Oracle variants are additionally provided ground-truth surface labels to analyze the impact of explicit mode information on robustness.

\section{Results}
\label{sec:results}

We evaluate Sim2Sim2Sim through a sequence of experiments assessing training and inference efficiency, open-loop dynamics quality, closed-loop policy learning in distilled dynamics, transferability to high-fidelity simulation, and the factors governing policy robustness across dynamics models and training regimes.

\subsection{Training and Inference Efficiency}
\label{sec:efficiency}

A key motivation for Sim2Sim2Sim is to enable scalable RL policy optimization without requiring direct training in high-fidelity simulators, which is computationally prohibitive due to the millions of environment steps typically required.

In practice, policy training in GPUDrive~\cite{Kazemkhani2024GPUDrive} achieves substantially higher throughput than approaches that rely on direct training in high fidelity simulators. 
Recent related works report wall-clock training times of several days when learning policies directly in high-fidelity physics simulators. For example, SPARC~\cite{Grooten2025SPARC} requires 6 days for 6M steps, and LocoFormer~\cite{Liu2025LocoFormer} reports around 7 days of training. 
We also attempted direct RL training in BeamNG using SAC~\cite{haarnoja2018SAC} with the same observation and reward formulation, but found that policies failed to converge within a comparable training budget, with BeamNG achieving only approximately 2 environment steps per second (SPS). 
In contrast, our framework achieves approximately 6{,}000 SPS within the distilled dynamics environment, reaching 150M environment steps in approximately 7 hours using PPO~\cite{Schulman2017PPO}---a roughly 3{,}000$\times$ improvement in training throughput.

Beyond training efficiency, the resulting policies also admit fast inference suitable for real time control. 
In BeamNG~\cite{beamng_tech}, RL policy inference requires less than 2 ms per step as seen in \cref{tab:tracking_and_latency_full_speed}, compared to tens of milliseconds typically required by MPC~\cite{patel2025recent}. 

\begin{table}[t]
    \centering
    \small
    \caption[Dynamics Model Evaluation]{Open-loop trajectory prediction errors for different dynamics models on the asphalt validation set.}
    \begin{tabular}{lccc}
        \hline
        \textbf{Model} & \textbf{SSDE (m)} & \textbf{ADE@10 (m)} & \textbf{FDE@10 (m)} \\
        \hline
        Bicycle & 0.035 & 0.259 & 0.537 \\
        DDM     & 0.010 & 0.128 & 0.334 \\
        Trans   & \textbf{0.007} & \textbf{0.089} & \textbf{0.206} \\
        \hline
    \end{tabular}
    \label{tab:dynamics_error}
\end{table}

\subsection{Open-loop Dynamics Learning Evaluation}
\label{sec:dynamics_results}
As shown in \cref{tab:dynamics_error}, we evaluate dynamics models trained on the asphalt subset (as described in \cref{sec:eval_protocol}) under open-loop rollout on a held-out asphalt validation set to characterize their short-horizon predictive behavior. 
We compare three representative model families—a kinematic bicycle model, DDM, and a fully data-driven Transformer.

Across all models, single-step errors are relatively small, while differences become more pronounced under multi-step rollout. 
The Transformer achieves the lowest open-loop error across all metrics, followed by the DDM, with the bicycle model performing worst. 
The DDM improves over the bicycle model by incorporating physical structure that better captures vehicle dynamics, while the Transformer further reduces open-loop error through increased model capacity and freedom. 
However, even the best-performing model accumulates non-negligible error over a short horizon, with final displacement errors on the order of 20–30 cm after 10 steps. 
Errors of this magnitude are already large enough to pose challenges for standard model-based control, particularly in settings with limited friction margins.

\begin{figure}[t]
\centering
\includegraphics[width=\linewidth]{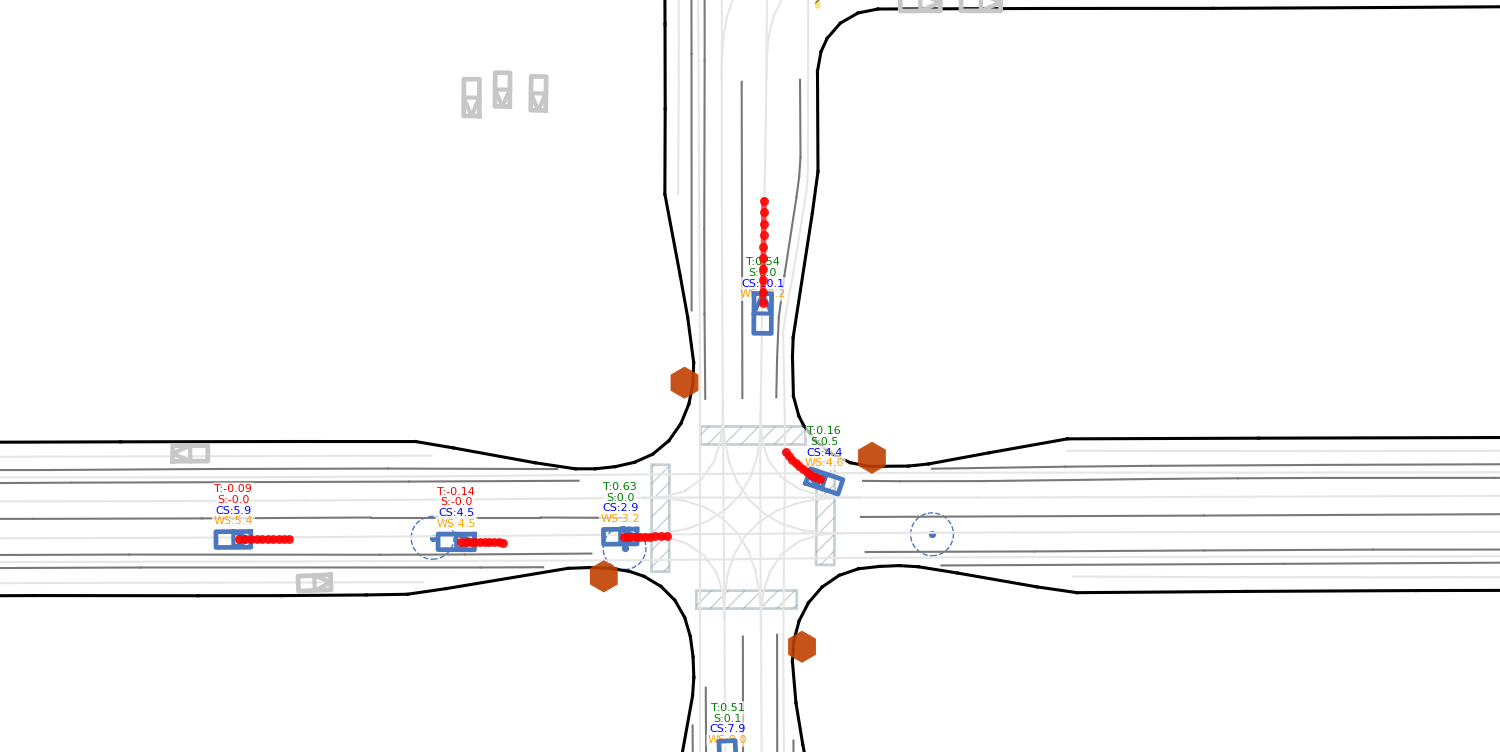}
\vspace{0.5em}
\includegraphics[width=\linewidth]{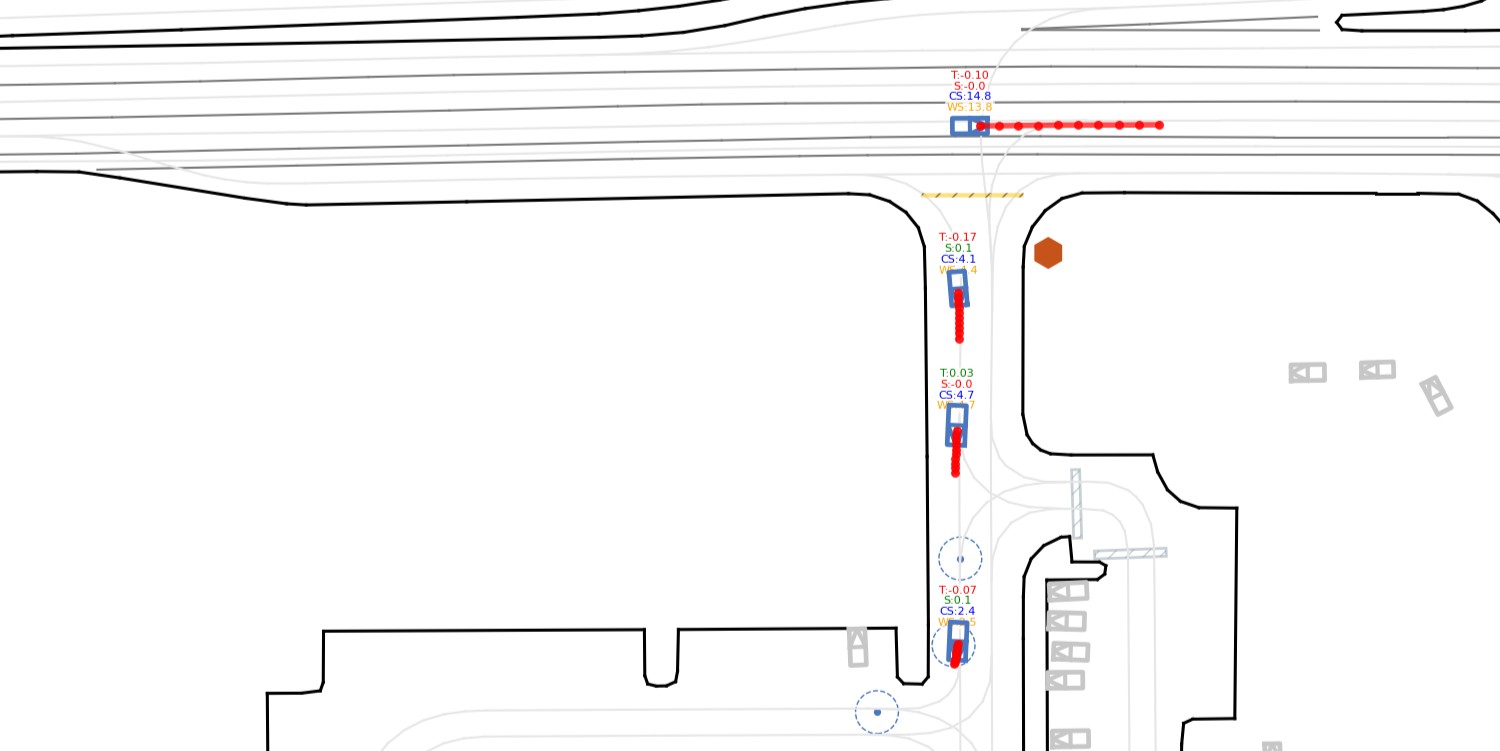}
\caption[WOMD driving scenarios]{Example driving scenarios from the WOMD Mini validation set. Red dots represent planned trajectories for each vehicle, with dashed circles indicating goal regions. \textbf{Top:} Vehicles navigating a T-intersection with mixed turning and straight-through maneuvers. \textbf{Bottom:} Slow-speed navigation on local roads with tight spacing. The diversity of traffic patterns, vehicle sizes, and interaction complexity naturally encourages policy generalization.}
\label{fig:womd_scenarios}
\end{figure}

\begin{table*}[t]
    \centering
    \small
    \setlength{\tabcolsep}{5.5pt}
    \caption{Closed-loop policy performance in GPUDrive on WOMD Mini Val under nominal asphalt conditions, with policies trained and evaluated within the same dynamics backend using full-state observations.} 
    \begin{tabular}{lccccccccc}
        \hline
        \textbf{Model} & \textbf{CTE (m)} & \textbf{ADE (m)} & \textbf{FDE (m)} & \textbf{PTE (m)} & \textbf{HE (rad)} & \textbf{STE (m/s)} & \textbf{Success (\%)} & \textbf{Collision (\%)} & \textbf{Off-road (\%)} \\
        \hline
        Bicycle & 0.27 & 1.20 & 1.85 & \textbf{1.59} & 0.04 & 0.42 & 91.72 & 0.87 & \textbf{1.65} \\
        DDM     & 0.30 & 1.27 & 1.96 & 1.65 & \textbf{0.03} & 0.42 & 92.07 & \textbf{0.54} & \textbf{1.65} \\
        Trans   & \textbf{0.19} & \textbf{1.14} & \textbf{1.62} & 1.95 & \textbf{0.03} & \textbf{0.40} & \textbf{94.42} & 1.67 & 2.12 \\
        \hline
    \end{tabular}
    \label{tab:closed_loop_results_full_state}
\end{table*}

\begin{table*}[t]
    \centering
    \small
    \caption{Closed-loop policy performance on WOMD E2E turning scenarios evaluated under icy conditions.} 
    \begin{tabular}{llccccccc}
        \hline
        \textbf{Model} & \textbf{Training Env} & \textbf{CTE (m)} & \textbf{ADE (m)} & \textbf{FDE (m)} & \textbf{PTE (m)} & \textbf{HE (rad)} & \textbf{STE (m/s)} & \textbf{Success (\%)} \\
        \hline
        Trans C        & Asphalt        & 0.99 & 2.32 & 7.87  & 2.29 & 0.16 & 0.45 & 27.50 \\
        \rowcolor{gray!15}
        Trans C        & Asphalt \& Ice & 0.59 & \textbf{1.87} & \textbf{5.92} & \textbf{1.84} & \textbf{0.15} & \textbf{0.43} & \textbf{39.00} \\
        \hline
        DDM + Re-C     & Asphalt        & \textbf{0.52} & 3.84 & 12.54 & 3.78 & 0.22 & 0.56 & 18.00 \\
        \rowcolor{gray!15}
        DDM + Re-C     & Asphalt \& Ice & \textbf{0.52} & 3.62 & 11.97 & 3.56 & 0.20 & 0.54 & 19.50 \\
        \hline
    \end{tabular}
    \label{tab:conditional_training_env}
\end{table*}

\begin{table*}[t]
    \centering
    \small
    \caption{Tracking performance, steering smoothness, and inference latency on the Putnam Park track under nominal asphalt conditions across different observation settings. All policies complete the full track. Results show mean~±~std over 10 evaluation runs.}
    \begin{tabular}{llccccccc}
        \hline
        \textbf{Model} &
        \textbf{Obs} &
        \textbf{CTE (m)} &
        \textbf{PTE (m)} &
        \textbf{HE (rad)} &
        \textbf{STE (m/s)} &
        \textbf{Steer D1 ($\mathbf{10^{-2}}$)} &
        \textbf{Inference (ms)} \\
        \hline
        Bicycle & Speed &
        0.26 ± 0.03 &
        0.98 ± 0.09 &
        0.05 ± 0.008 &
        0.92 ± 0.11 &
        6.28 ± 0.71 &
        1.90 ± 0.11 \\
        \rowcolor{gray!15}
        Bicycle & Full &
        \textbf{0.09 ± 0.01} &
        0.99 ± 0.08 &
        \textbf{0.01 ± 0.002} &
        \textbf{0.24 ± 0.03} &
        1.79 ± 0.21 &
        1.88 ± 0.12 \\
        \hline
        DDM & Speed &
        0.10 ± 0.01 &
        0.92 ± 0.07 &
        0.01 ± 0.003 &
        0.30 ± 0.04 &
        1.46 ± 0.19 &
        \textbf{1.83 ± 0.11} \\
        \rowcolor{gray!15}
        DDM & Full &
        \textbf{0.09 ± 0.01} &
        \textbf{0.89 ± 0.06} &
        \textbf{0.01 ± 0.003} &
        0.38 ± 0.04 &
        \textbf{0.73 ± 0.11} &
        1.84 ± 0.10 \\
        \hline
        Trans & Speed &
        0.50 ± 0.05 &
        1.09 ± 0.10 &
        0.05 ± 0.007 &
        2.25 ± 0.23 &
        7.06 ± 0.82 &
        1.91 ± 0.13 \\
        \rowcolor{gray!15}
        Trans & Full &
        \textbf{0.09 ± 0.02} &
        0.95 ± 0.07 &
        0.02 ± 0.004 &
        0.30 ± 0.05 &
        3.37 ± 0.45 &
        1.90 ± 0.14 \\
        \hline
    \end{tabular}
    \label{tab:tracking_and_latency_full_speed}
\end{table*}

\subsection{Policy Learning and Evaluation in Distilled Dynamics}
\label{sec:policy_learning}

We next evaluate policy learning performance within GPUDrive using different learned dynamics models as the simulation backend (\cref{fig:womd_scenarios}). 
As shown in Table~\ref{tab:closed_loop_results_full_state}, all policy--dynamics pairs achieve high success rates above 90\%, indicating sufficient compatibility for basic policy learning across all dynamics backends. 
Success rates vary modestly (91.72\%--94.42\%), but more pronounced differences emerge in tracking accuracy and failure modes: the Transformer-based dynamics yields the lowest cross-track error and highest success rate, yet also the highest collision and off-road rates. This suggests that despite its expressive capacity, the Transformer dynamics model fails to capture the physical nuances necessary for the policy to learn safe maneuvering behavior, even within its own training environment.

To further stress-test the policies, we evaluate them on turning scenarios extracted from the Waymo E2E dataset~\cite{xu2025wod} with surface transitions introduced into these scenes, where icy conditions are applied from timestep 3~s onward.
Turning maneuvers amplify lateral dynamics and friction sensitivity, providing a more demanding testbed than straight-line icy driving. 
As shown in \cref{tab:conditional_training_env}, policies trained with exposure to ice consistently outperform those trained on asphalt only when encountering abrupt friction transitions.

For DDM-based models, we introduce a learned residual module conditioned on surface type to capture surface-dependent deviations that cannot be expressed by the base dynamics alone. 
Adding conditional residual on the DDM yields limited behavioral change, as the strong inductive bias imposed by the underlying bicycle dynamics constrains the range of admissible responses, leading to only limited performance differences. 
We defer a deeper analysis of these interactions to \cref{sec:policy_robustness}.

\subsection{Transfer to High-Fidelity Simulation}
\label{sec:transfer_results}

\begin{figure}[t]
\centering
\includegraphics[width=\linewidth]{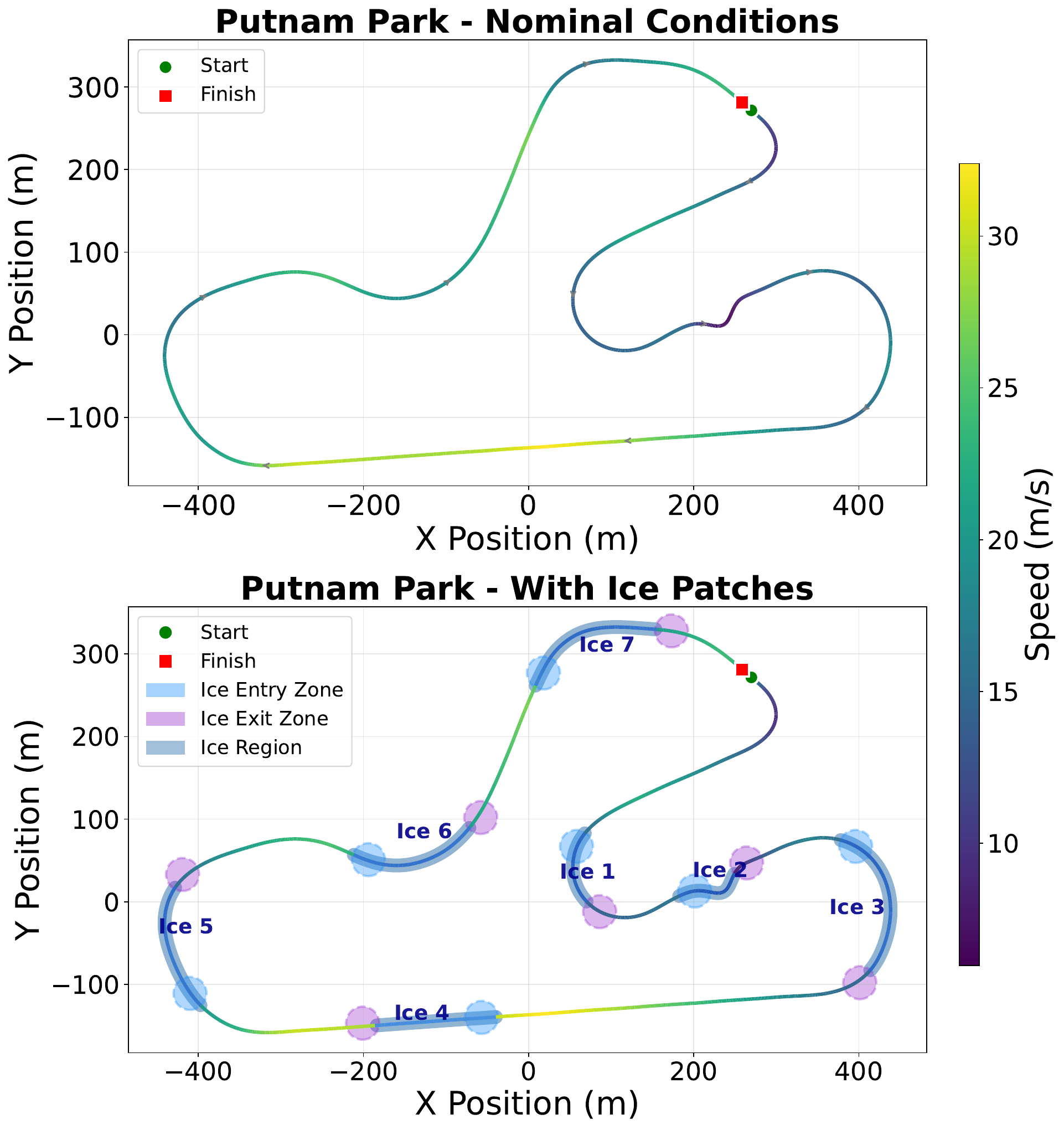}
\caption[Putnam Park Road Course evaluation environments]{Evaluation tracks in BeamNG. \textbf{Top:} Putnam Park Road Course (2.765 km) under nominal asphalt conditions. \textbf{Bottom:} same track modified with seven ice patches (blue regions) creating friction transitions, with marked entry/exit zones. Ice patches test the policies' robustness to sudden dynamics changes where policies must rapidly adapt their control strategy.}
\label{fig:putnam}
\end{figure}

We evaluate the transfer performance of policies trained in GPUDrive by deploying them back into the high-fidelity BeamNG simulator without any fine-tuning. 
Policies are tested on racecar trajectories collected from the Putnam Park Road Course~\cite{Kulkarni2023}, using the fastest lap as the reference trajectory (\cref{fig:putnam}). 
This setting provides a challenging evaluation scenario characterized by sustained high speeds and tight maneuvers.
As shown in \cref{tab:tracking_and_latency_full_speed}, all policies successfully complete the track and exhibit comparable inference latency below 2~ms; the policy trained with DDM as the dynamics backend achieves the best overall performance, with the lowest tracking error and the smoothest steering behavior.

To further assess transfer under more challenging conditions, we evaluate policies on the Putnam track with manually introduced ice patches (\cref{fig:putnam}). 
Under this setting, failure is defined by loss of vehicle stability leading to spin-out, making distance traveled and heading error the primary indicators of performance. 
As summarized in \cref{tab:putnam_ice_patch}, policies trained with exposure to ice during distilled-dynamics training consistently travel farther and exhibit reduced heading error. 
In particular, the DDM with residual conditioning trained on mixed asphalt and ice achieves the greatest tracking distance, completing 83.4\% of the track and exceeding that of the conditional Transformer.

\begin{table}[t]
    \centering
    \small
    \caption{Tracking stability and distance traveled on the Putnam Park track with ice 
    patches. Results show mean~±~std over 10 evaluation runs.}
    \begin{tabular}{llcc}
        \hline
        \textbf{Model} & \textbf{Training Env} & \textbf{HE (rad)} & \textbf{Dist. (\%)} \\
        \hline
        DDM & Asphalt &
        0.371 ± 0.249 & 29.6 ± 26.1 \\
        \hline
        DDM + Re-C & Asphalt &
        0.217 ± 0.198 & 40.4 ± 22.3 \\
        \rowcolor{gray!15}
        DDM + Re-C & Asphalt + Ice &
        0.189 ± 0.185 & \textbf{83.4 ± 15.4} \\
        \hline
        Trans C & Asphalt &
        0.319 ± 0.241 & 19.0 ± 18.5 \\
        \rowcolor{gray!15}
        Trans C & Asphalt + Ice &
        \textbf{0.068 ± 0.030} & 58.7 ± 12.7 \\
        \hline
    \end{tabular}
    \label{tab:putnam_ice_patch}
\end{table}

\subsection{Policy Robustness Emergence}
\label{sec:policy_robustness}

We demonstrate that policies trained entirely in distilled dynamics can handle abrupt dynamics transitions in high-fidelity simulation without any fine-tuning.
This section will present a detailed analysis of the factors governing policy's robustness.

\textbf{Richer Observations Modulate Policy--Dynamics Interaction.}
As shown in \cref{tab:tracking_and_latency_full_speed}, restricting observations to scalar speed leads to pronounced degradation across all dynamics backends: cross-track error increases by up to 456\% (0.09$\to$0.50) and heading error by up to 400\% (0.01$\to$0.05) for bicycle- and Transformer-based policies. 
This degradation arises because a single speed scalar entangles longitudinal and lateral motion, preventing the policy from distinguishing forward progress from lateral slip in realistic BeamNG dynamics. 
Providing $v_x$, $v_y$, and $\omega$ restores this separation, enabling stable tracking even when slip is not explicitly modeled in the simpler bicycle dynamics backend. DDM-based policies are more resilient to this reduction, reflecting the benifites of stronger physical inductive bias.

\textbf{Open-Loop Dynamics Model Accuracy Is Insufficient for Robust Control Learning.}
Although the Transformer achieves the lowest open-loop prediction error (\cref{tab:dynamics_error}) and maintains this advantage in distilled-dynamics policy evaluation (\cref{tab:closed_loop_results_full_state}), this ranking does not persist after simulator transfer.
As shown in \cref{tab:tracking_and_latency_full_speed}, policies trained in DDM-based environments exhibit lower heading error and smoother steering in BeamNG than those trained with a Transformer-based backend, with even the simple bicycle model producing smoother control than the Transformer despite its inferior open-loop predictive accuracy.
One possible explanation is that highly expressive dynamics models allow the RL policy to exploit subtle modeling artifacts during training, leading to behaviors that are less robust under distribution shift in high-fidelity simulation.
In such cases, iterative data aggregation strategies (e.g., DAgger-style retraining of the dynamics model using policy-induced rollouts~\cite{ross2011dagger}) may reduce model bias in off-distribution regions and improve transfer robustness.
Dynamics models should therefore be evaluated not only by predictive accuracy, but also by how they shape closed-loop policy behavior and support stable deployment after transfer to high-fidelity environments.

\textbf{Physics Structure Alone Is Insufficient for Out-of-Distribution Robustness.}
As shown in \cref{tab:putnam_ice_patch}, although the DDM-trained policy achieves the strongest baseline on nominal asphalt in \cref{tab:tracking_and_latency_full_speed} due to its strong physical inductive bias, it still fails under friction shifts, traveling only 29.6\% of the track before losing control and exhibiting the highest heading error. 

To isolate the effect of training diversity, we compare surface-conditional models trained on asphalt-only versus mixed asphalt-and-ice environments. Removing ice from the training distribution leads to substantial degradation for both DDM + Re-C and Trans C, reducing distance traveled by over 50\% (83.4\%$\to$40.4\%) and 65\% (58.7\%$\to$19.0\%) respectively, demonstrating that even a strong physical inductive bias is no substitute for sufficient exposure to out-of-distribution conditions during training.
Without sufficient exposure to out-of-distribution conditions, structural advantages alone cannot sustain robust behavior at deployment.
This robustness emerges purely from training diversity rather than architectural adaptation: unlike prior approaches such as RAPTOR~\cite{Eschmann2025RAPTOR}, LocoFormer~\cite{Liu2025LocoFormer}, and SPARC~\cite{Grooten2025SPARC}, which rely on history or context encoders, our policies observe only the current vehicle state---suggesting that explicit history-based adaptation may not be strictly necessary when the training distribution is sufficiently rich.

\textbf{Robust Control Without Explicit Surface Labels.}
As seen in \cref{tab:label_obs_ablation}, policies trained on mixed asphalt and ice data without surface labels in the observation (non-oracle) consistently outperform their label-conditioned counterparts under icy evaluation in the high-fidelity BeamNG simulator.
These results indicate that explicit mode labels do not necessarily improve robustness when the dynamics model is imperfect: conditioning on surface labels can cause the policy to rely on potentially mismatched model-conditioned behavior during training, which does not reflect the real dynamics. 
In contrast, policies trained without surface labels must rely on physical feedback such as lateral velocity and yaw rate, leading to more conservative and robust behavior under friction changes.

In summary, policy robustness arises from the joint effects of dynamics fidelity, observation design, and training diversity. Additionally, these findings suggest that a dynamics model should be assessed not only by its predictive accuracy alone, but also by the quality of the policies it enables.

\begin{table}[t]
    \centering
    \small
    \setlength{\tabcolsep}{3.3pt}
    \caption{Effect of providing ground-truth surface labels as part of the policy observation under ice-patch evaluation on the Putnam Park track.}
    \begin{tabular}{lcccc}
        \hline
        \textbf{Model} & \textbf{Label} & \textbf{HE (rad)} & \textbf{Dist. (\%)} & 
        \makecell{\textbf{Steer D1} \\ \textbf{($10^{-2}$)}} \\
        \hline
        DDM + Re-C & Yes &
        0.235 ± 0.201 & 28.8 ± 19.3 & \textbf{1.15 ± 0.18} \\
        \rowcolor{gray!15}
        DDM + Re-C & No &
        0.189 ± 0.185 & \textbf{83.4 ± 15.4} & 1.22 ± 0.14 \\
        \hline
        Trans C & Yes &
        0.246 ± 0.187 & 53.5 ± 14.2 & 2.03 ± 0.21 \\
        \rowcolor{gray!15}
        Trans C & No &
        \textbf{0.068 ± 0.030} & 58.7 ± 12.7 & 2.42 ± 0.19 \\
        \hline
    \end{tabular}
    \label{tab:label_obs_ablation}
\end{table}

\section{Conclusion}
\label{sec:conclusion}
We presented Sim2Sim2Sim, a framework that bridges high-fidelity vehicle simulation and scalable reinforcement learning for robust autonomous driving control. 
Through extensive closed-loop evaluation, we demonstrated that policies trained on distilled dynamics can transfer reliably to high-fidelity simulation to handle challenging friction transitions, and that open-loop predictive accuracy alone is insufficient to characterize a dynamics model's suitability as a training environment for RL control policy learning. 
These findings highlight dynamics distillation as a practical pathway toward scalable and reliable policy learning. An exciting future direction involves investigating if more expressive dynamics models, such as autoregressive world models~\cite{Li2025RWM}, can improve distillation fidelity which leads to more robust, generalizable control policies.

{
    \small
    \bibliographystyle{ieeetr}
    \bibliography{main}
}

\end{document}